\documentclass[conference]{IEEEtran}
\IEEEoverridecommandlockouts
\usepackage{cite}
\usepackage{enumitem}
\usepackage{amsmath,amssymb,amsfonts}
\usepackage{algorithmic}
\usepackage{graphicx}
\usepackage{textcomp}
\usepackage[flushleft]{threeparttable}
\usepackage{xcolor}
\usepackage{soul}
\usepackage{url}
\usepackage[utf8]{inputenc}
\usepackage{color, colortbl}
\usepackage{adjustbox}
\usepackage{subcaption}
\usepackage{booktabs,caption}
\usepackage{hhline, multirow, makecell}
\usepackage{amsthm}
\usepackage{booktabs}
\usepackage{algorithm}
\newcommand{\norm}[1]{\left\lVert#1\right\rVert}
\def\BibTeX{{\rm B\kern-.05em{\sc i\kern-.025em b}\kern-.08em
    T\kern-.1667em\lower.7ex\hbox{E}\kern-.125emX}}
\begin{document}

\title{An Adversarial Domain Separation Framework for Septic Shock Early Prediction Across
EHR Systems 
}

\author{\IEEEauthorblockN{Farzaneh Khoshnevisan\IEEEauthorrefmark{1} and Min Chi\IEEEauthorrefmark{2}}
\IEEEauthorblockA{
\textit{Department of Computer Science} \\
 North Carolina State University, Raleigh, USA \\
\{fkhoshn\IEEEauthorrefmark{1}, mchi\IEEEauthorrefmark{2}\}@ncsu.edu}
}

\IEEEoverridecommandlockouts
\IEEEpubid{\makebox[\columnwidth]{978-1-7281-6251-5/20/\$31.00~\copyright2020 IEEE \hfill}
\hspace{\columnsep}\makebox[\columnwidth]{ }}

\maketitle
\IEEEpubidadjcol

\begin{abstract}
Modeling patient disease progression using Electronic Health Records (EHRs) is critical to assist clinical decision making. While most of prior work has mainly focused on developing effective disease progression models using EHRs collected from \emph{an individual} medical system, relatively little work has investigated building robust yet generalizable diagnosis models \emph{across different systems}.  In this work, we propose a general domain adaptation (DA) framework that tackles two categories of discrepancies in EHRs collected from different medical systems: one is caused by heterogeneous patient populations (covariate shift) and the other is caused by variations in data collection procedures (systematic bias). Prior research in DA has mainly focused on  addressing covariate shift but not systematic bias. In this work, we propose an adversarial domain separation framework that addresses both categories of discrepancies by maintaining  \emph{one globally-shared} invariant latent representation \emph{across all systems} through an adversarial learning process, while also allocating a \emph{domain-specific} model for \emph{each system} to extract local latent representations that cannot and should not be unified across systems. Moreover, our proposed framework is based on variational recurrent neural network (VRNN) because of its ability to capture complex temporal dependencies and handling missing values in time-series data. We evaluate our framework for early diagnosis of an extremely challenging condition, septic shock, using two real-world EHRs from distinct medical systems in the U.S. The results show that by \emph{separating globally-shared from domain-specific representations}, our framework significantly improves septic shock early prediction performance in both EHRs and outperforms the current state-of-the-art DA models.

\end{abstract}


\section{Introduction}\label{sec:intro}
Electronic Health Records (EHRs) are large-scale and systematic collection of temporal health information of patients. The broad adoption of EHRs in medical systems has promoted the development of various computational methods for understanding the medical history of patients and predicting risks. A variety of classical and deep machine learning approaches have been applied to EHRs for clinical outcome prediction. While majority of past studies develop highly effective algorithms for EHRs collected from \emph{an individual medical system}, relatively little work has investigated development of robust yet generalizable prediction models \emph{across different medical systems}.  Given that different medical systems often serve populations with different demographic information, employ different infrastructure, and often adopt different workflows and administrative policies \cite{agniel2018biases},  EHRs across medical systems can vary dramatically. For the purpose of this work, we refer the discrepancies caused by the heterogeneous patient populations across EHR systems as \emph{covariate shift} and those caused by incompatible data collection procedures as \emph{systematic bias}. 
We compared two large-scale EHRs collected from July, 2013 to December, 2015 in two medical systems in the U.S.:  Christiana Care Health System (CCHS) located in Newark, Delaware and Mayo Clinic located in Rochester, Minnesota. For covariate shift, our analysis shows that the two systems' population demographics are very different due to distinct hospital locations; for example, African American patients constitute 22.5\% of the whole population in CCHS, while they constitute only about 2\% of Mayo population. As a result, most clinical variables have divergent marginal distributions across the two EHRs. For example, the Kullback-Leibler divergence between Bilirubin measurement distribution in CCHS vs. Mayo is 1.39. As for systematic bias, CCHS and Mayo follow vastly different procedures for collecting EHRs; for example, vital signs and lab values are measured every 44 minutes and 38.1 hours on average in Mayo but they are measured every 244 minutes and 47.8 hours in CCHS, respectively. As a result, directly utilizing models that are trained on one system to another system, or training prediction models using combined EHRs across systems often fails to result in robust yet generalizable models.

In this work, we propose a general adversarial domain adaptation (DA) framework that tackles both covariate shift and systematic bias across different EHR systems. Prior work has mainly focused on either adapting a prediction model that is pre-trained on one system by fine-tuning its parameters to the new one \cite{alves2018dynamic}, or learning an invariant representation that is shared between two systems, such as VRADA \cite{purushotham2016varda}. While such approaches can be effective for handling covariate shift, they may not be effective for systematic bias because the latter is often caused by different administrative policies or infrastructures of different medical systems that cannot and should not be unified across systems. Our proposed framework, in contrast, maintains \emph{one globally-shared} invariant latent representation across \emph{all systems} through an adversarial learning process to handle covariate shift, while it also allocates a \emph{domain-specific} model for each system to extract local latent representations that should not be unified across systems and handles systematic bias. Furthermore, unlike previous DA work that treats source and target domains differently, the symmetric architecture of this framework does not differentiate between them and extends its application to multi-domain adaptation and domain generalization scenarios.

We evaluate our proposed framework for early detection of an extremely challenging condition in hospitals, septic shock. Sepsis is a life-threatening organ dysfunction caused by a dysregulated body response to infection \cite{singer2016third}. Sepsis is the third leading cause of in-hospital mortality and is among the most expensive conditions treated in the U.S. \cite{cdc2016,torio2006national}. Septic shock, the most severe stage of sepsis, has 40-70\% mortality rate. However, as much as 80\% of deaths can be prevented through early detection and rapid treatment with antibiotics \cite{kumar2006duration}. 
 Timing is essential as every hour of delay in treatment leads to an 8\% increase in mortality for patients who exhibit septic shock. 
\emph{Early prediction of sepsis is difficult} since the early symptoms of sepsis are typically vague and include subtle changes in mental status or minor changes in white blood-cell count \cite{kumar2006duration}. 
A diverse set of tools and guidelines for the early diagnosis and treatment of sepsis have been developed by clinical researchers. For example, multiple bedside scoring systems, such as SOFA, qSOFA, APACHE II, and PIRO score, are widely used in hospitals for detecting at-risk septic patients \cite{macdonald2014comparison}. Unfortunately these general scoring systems often lack the necessary sensitivity and specificity for identifying high-risk septic shock patients \cite{henry2015targeted,dorsett2017qsofa,mak2019prospective}. This is because sepsis, like cancer, involves various disease etiologies that span a wide range of syndromes, and different patient groups may show vastly different symptoms \cite{tintinalli2015sepsis}. Therefore, a general yet robust diagnosis model that can predict septic shock several hours before the onset is in high demand.  

Prior work, including some of our own work,  has explored deep recurrent neural network (RNN) and its variations especially long short-term memory (LSTM), for the task of early prediction of septic shock \cite{lin2019multi,lin2018early,zhang2017lstm,kim2018temporal,khoshnevisan2018recent}.  While LSTMs can capture long-range temporal dependencies in EHRs and have shown great success in this task,  variational recurrent neural network (VRNN) \cite{chung2015recurrent} has gained its popularity for sequential EHR data modeling because of its ability to handle variabilities in EHRs, such as missing data, and its ability to capture complex conditional and temporal dependencies \cite{zhang2017medical,mulyadi2020uncertainty}. In this work,  the effectiveness of VRNN was compared to LSTM for early septic shock prediction tasks and our results show that  VRNN significantly outperforms LSTM on every measure. As a result, our proposed framework is VRNN-based.

In short, we propose a  VRNN-based DA framework that  \emph{separates} a global latent representation across all systems from domain-specific representations for each system and thus we refer to it as \textbf{\emph{VRNN-based Adversarial Domain Separation (VR-ADS)}}. 
The effectiveness of our proposed VR-ADS framework is validated on real-world EHRs from two medical systems, CCHS and Mayo, for the challenging task of septic shock early prediction up to 48 hours before the onset. We explicitly demonstrate that how addressing both covariate shift and systematic bias embedded in our framework improves predictions across both EHR systems.
Overall, our results show that VR-ADS framework has at least the following two main contributions:
\begin{itemize}
    \item Our proposed framework significantly outperforms non-adaptive baselines as well as state-of-the-art DA approaches, including Fine-tuning \cite{alves2018dynamic} and VRADA \cite{purushotham2016varda}. More importantly, we demonstrate that the effectiveness of VR-ADS indeed comes from the fact that it separates the \emph{global} representation from the \emph{local} ones to address systematic bias. We show this by introducing a DA framework that only unifies latent representations, referred to as adversarial domain unification (ADU), and show that it does not perform as good as VR-ADS.
    \item Our proposed framework can be used to address insufficient labeled data problem. We show that by combining training data across different EHR systems, the performance of VR-ADS in detecting septic patients 48 hours before the onset is improved by more than 9\%, compared with models trained on each dataset separately.
\end{itemize}

\section{related Work}
\noindent{\textbf{Septic Shock Early Prediction:}}  
Prior research has investigated the application of various traditional and deep machine learning models for septic shock early prediction. A number of classical and statistical approaches, such as multivariate logistic regression and cox proportional hazards, have been developed as early warning systems for risk prediction several hours before the septic shock onset \cite{shavdia2007septic,henry2015targeted}. Moreover, sequential pattern mining approaches in conjunction with classic learning algorithms are employed by previous studies for early prediction of septic shock \cite{ghosh2017septic,khoshnevisan2018recent,mao2019one}. Such models have shown to be not only effective but also produce explainable patterns for this task. In recent years, various deep learning models have been applied to EHRs for septic shock early prediction. Among them, variations of recurrent neural network (RNN), especially LSTM, have gained a lot of attention \cite{zhang2017lstm,lin2018early,zhang2019attain}. Despite the great power of LSTM-based models, they are not designed to directly address missing values in EHR \cite{kim2018temporal,yang2018time}. Variational recurrent neural network (VRNN) \cite{chung2015recurrent} is recently proposed to model complex temporal and conditional dependencies in sequential data as it has shown great success in speech modeling and NLP \cite{chien2017variational,pineau2019variational}. Furthermore, VRNN is composed of a generative process that enables it to impute missing values in time-series data and makes it a great fit for modeling sequential EHR data as previous studies have also shown \cite{zhang2017medical,mulyadi2020uncertainty}. As far as we know, this is the first work that applies VRNN for septic shock early prediction.

\noindent \textbf{Domain adaptation (DA)} is a sub-category of transfer learning in which the feature spaces between source and target domains are shared while their marginal distributions differ. Generally, DA models attempt to either learn a mapping from one domain to another and unify domains \cite{tzeng2017adda}, or find a common space between domains to generate domain-invariant representations \cite{pan2010domain,long2015learning,tzeng2015simultaneous}. As a result, most of the existing models address the covariate shifts between the source and target domains regardless of systematic bias, and leave the shared representations vulnerable to contamination. 
Classic DA models address the covariate shift either using instance re-weighting to make source samples more similar to target \cite{yao2010boosting}, or using feature alignment to align source feature space to target, usually using kernel-based models \cite{fernando2013unsupervised}.

Among deep DA approaches, network-based adaptation and feature transformation have became popular in recent years \cite{wang2018deep}. \emph{Network-based DA} pre-trains a deep network on source domain and shares its parameters with target network for fine-tuning.
Alvez et al. compared five different network-based DA variations in a CNN-LSTM framework for mortality prediction using EHR and showed that the fine-tuning model outperformed all others \cite{alves2018dynamic}. As a result, in this work we use a fine-tuning approach as a baseline. \emph{Feature transformation DA} approaches project both source and target domains into a common latent space to minimize the covariate shift. With recent advancements in Generative Adversarial Network (GAN) \cite{gan2014goodfellow}, great progress has been made in employment of adversarial learning for DA and they have shown great success, specially in computer vision and NLP \cite{tzeng2017adda,hoffman2017cycada}. Discriminative adversarial DA models leverage the discriminator's power to align data distribution across domains and they can be applied to various types of data \cite{ganin2014unsupervised,shen2018wasserstein}.
Two previous studies have employed adversarial learning for patient sub-population adaption using sequential EHR data. T-ALSTM \cite{zhang2019time} is designed for event-level adaptation across different age, race, and gender groups for prediction of septic shock, but its architecture is not compatible with our visit-level prediction task. VRADA \cite{purushotham2016varda} is a VRNN-based adversarial DA framework that its architecture is the closest to ours and thus, in the following we include it as a baseline as well. 

\noindent{\textbf{Cross-hospital DA}} models in literature are mostly based on the traditional machine learning models that are not designed to handle temporal and non-linear dependencies in EHR. Further, most studies focus on either feature space matching \cite{wiens2014study} or addressing covariate shift using approaches such as instance re-weighting \cite{desautels2017using,curth2019transferring}. Recently, Mhasawade et al. proposed a hierarchical Bayesian DA model as a sub-population adaptation task to capture information shared among subgroups of patients across hospitals \cite{mhasawade2019population}.
To the best of our knowledge, no prior DA work has addressed incompatible data collection procedure or systematic bias across medical systems. 

\section{Problem Setup}

For the purpose of this paper, we assume that our task involves two domains: $\mathcal{D}_1$  and $\mathcal{D}_2$. Each domain contains a certain number of hospital visits represented as $\boldsymbol{X}=\{ \boldsymbol{\mathrm{x}}^1,...,\boldsymbol{\mathrm{x}}^{n}\}$ where $n$ is the number of visit. Each visit $\boldsymbol{\boldsymbol{\boldsymbol{\mathrm{x}}}}^i$ is a multivariate time-series that is composed of $T^i$ medical events and can be denoted as $\boldsymbol{\mathrm{x}}^i=(\boldsymbol{x}_t^i)_{t=1}^{T^i}$ where $\boldsymbol{x}_t^i \in \mathbb{R}^D$. Additionally, for each visit,  it also has a visit-level outcome label represented as $\boldsymbol{Y}=\{ \mathrm{y}^1,...,\mathrm{y}^{n}\}$ where $\mathrm{y^i}\in\{1,0\}$ indicates the outcome of visit $i$:  septic shock vs.  non-septic shock.
By combining $\boldsymbol{X}$ and $\boldsymbol{Y}$ for each domain, we have:  $\mathcal{D}_1=\{\boldsymbol{\mathrm{x}}_{\mathcal{D}_1}^i,\mathrm{y}_{\mathcal{D}_1}^i\}_{i=1}^{n_1}$  and $\mathcal{D}_2=\{\boldsymbol{\mathrm{x}}_{\mathcal{D}_2}^j,\mathrm{y}_{\mathcal{D}_2}^j\}_{j=1}^{n_2}$,  where $n_1$ and $n_2$ are the number of visits in $\mathcal{D}_1$ and $\mathcal{D}_2$ respectively; In our problem settings, we assume that $\{\boldsymbol{\mathrm{x}}_{\mathcal{D}_1}^i,\mathrm{y}_{\mathcal{D}_1}^i\}_{i=1}^{n_1}$ is drawn from distribution $p_1(\boldsymbol{\mathrm{x}},\mathrm{y})$ while $\{\boldsymbol{\mathrm{x}}_{\mathcal{D}_2}^j,\mathrm{y}_{\mathcal{D}_2}^j\}_{j=1}^{n_2}$ is drawn from a different distribution $p_2(\boldsymbol{\mathrm{x}},\mathrm{y})$. Our objective is to minimize the discrepancies between these two domains in a common latent space by aligning their latent representations: $\boldsymbol{\mathrm{z}}_{\mathcal{D}_1}$ and $\boldsymbol{\mathrm{z}}_{\mathcal{D}_2}$, so that to create a unified, generalizable classifier $C : \boldsymbol{\mathrm{z}} \mapsto \mathrm{y}$ that predicts the outcome optimally in both domains. To do so, we apply adversarial learning to minimize the distance between $\boldsymbol{\mathrm{z}}_{\mathcal{D}_1}$ and $\boldsymbol{\mathrm{z}}_{\mathcal{D}_2}$ in two different DA frameworks. 

In the following, we will first introduce VRNN as the base classifier in Section \ref{vrnn} 
and then describe how adversarial learning in a GAN architecture helps adaptation procedure in Section \ref{sec:gan}. In Section \ref{sec:adu}, we introduce Adversarial Domain Unification (ADU) to compare against our VR-ADS framework proposed in Section \ref{sec:ads}.

\subsection{Variational Recurrent Neural Network (VRNN)}\label{vrnn}
 In general,  VRNN \cite{chung2015recurrent} extends the idea of variational auto-encoder (VAE) to a recurrent framework to capture dependencies between latent random variables across time steps. Thus, like VAE, VRNN has an encoder-decoder structure.

\noindent\textbf{Encoder:}
The recurrent criteria in VRNN is enforced by conditioning VAE at each time step by previous hidden state variable $\boldsymbol{h}_{t-1}$ of an RNN, such as an LSTM. Therefore, the prior follows the distribution: 
\begin{equation}
    \boldsymbol{z}_t \sim \mathcal{N}(\mu_{0,t}, \text{diag} (\sigma^2_{0,t})) \text{, where } [\mu_{0,t},\sigma_{0,t}] = \varphi^{\text{prior}}_{\tau}(\boldsymbol{h}_{t-1})
\end{equation}

\noindent and $\mu_{0,t}$, $\sigma_{0,t}$ denote parameters of the conditional prior distribution. Further, for each time step $\boldsymbol{x}_t$, latent variable $\boldsymbol{z}_t$ is inferred via:  
\begin{gather}
\begin{gathered}
    \boldsymbol{z}_t|\boldsymbol{x}_t \sim \mathcal{N}(\mu_{z,t}, \text{diag} (\sigma^2_{z,t})),\\ 
    \text{ where } [\mu_{0,t},\sigma_{0,t}] = \varphi^{\text{enc}}_{\tau}(\varphi^x_\tau(\boldsymbol{x}_t),\boldsymbol{h}_{t-1})
\end{gathered}
\end{gather}

\noindent and $\mu_{z,t}$, $\sigma_{z,t}$ are the parameters of the approximate posterior.  RNN updates its hidden state variable $\boldsymbol{h}_t$ recurrently using:
\begin{equation}
    \boldsymbol{h}_t = \varphi^{\text{rec}}_{\tau}([\varphi^x_\tau(\boldsymbol{x}_t), \varphi^z_\tau(\boldsymbol{z}_t)], \boldsymbol{h}_{t-1}).
\end{equation}

Additionally, $\varphi^{\text{prior}}_{\tau}$ and $\varphi^{\text{enc}}_{\tau}$ are approximated through deep neural networks, $\varphi^{x}_{\tau}$ and $\varphi^{z}_{\tau}$ are embedding networks of $\boldsymbol{x}_t$ and $\boldsymbol{z}_t$, and $\varphi^{\text{rec}}_{\tau}$ is an LSTM-based state transition function.

\noindent\textbf{Decoder:}
The generative model is conditioned on the latent state $\boldsymbol{z}_t$ and the previous hidden state $h_{t-1}$ and thus we have: 
\begin{equation}
\begin{gathered}
    \boldsymbol{x}_t|\boldsymbol{z}_t \sim \mathcal{N}(\mu_{x,t}, \text{diag} (\sigma^2_{x,t})),\\ \text{ where }
    [\mu_{x,t},\sigma_{x,t}] = \varphi^{\text{dec}}_{\tau}(\varphi^z_\tau(\boldsymbol{z}_t),\boldsymbol{h}_{t-1}),
\end{gathered}
\end{equation}

\noindent $\mu_{x,t}$, $\sigma_{x,t}$ are the parameters of the generating distribution, and $\varphi^{\text{dec}}_{\tau}$ is approximated by a deep neural network. 

\begin{figure}[tb]
\centering
\includegraphics[width=0.5\textwidth]{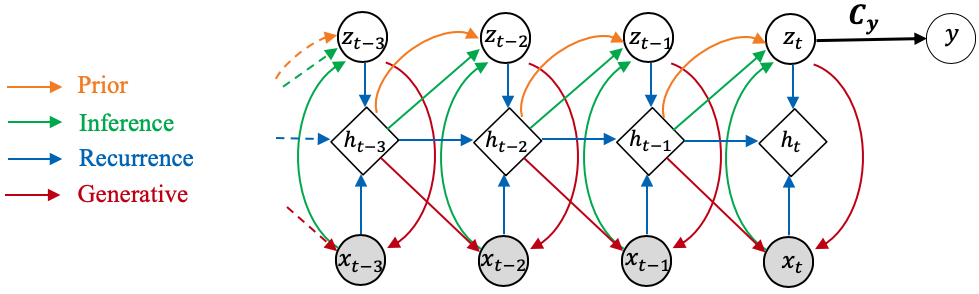}
\caption{\small{Sequential Diagram of VRNN Classifier}\label{fig:seq_vrnn}}
\end{figure}

\noindent\textbf{Classifier:}
A one-layer fully connected neural network is used for the classification task. It takes latent variable of the last step as input to capture dependencies across the time-steps and infers a label for the sequence. Therefore, for each $\boldsymbol{x}$, the input $\tilde{\boldsymbol{z}}_{T}$ is fed to the classifier such that:
\begin{equation}
    \tilde{\boldsymbol{z}}_{T} \sim q_{\theta_e}(\boldsymbol{z}_{T} | \boldsymbol{x}_{\leq T}, \boldsymbol{z}_{< T})
\end{equation}
\noindent where $q_{\theta_e}$ is the inference (encoder) model. Fig. \ref{fig:seq_vrnn} illustrates the recurrent structure of a VRNN classifier where four described processes are shown in different colors.

\noindent\textbf{Learning:}
The entire VRNN framework is learned by optimizing the encoder and decoder parameters, $\theta_e$ and $\theta_d$. The following objective function, $\mathcal{L}_\text{vrnn}(\boldsymbol{\mathrm{\boldsymbol{x}}};\theta_e,\theta_d)$, is derived by variational lower bound (ELBO) to minimize the reconstruction loss:

\begin{equation}
\begin{small}
\begin{aligned}
    \mathbb{E}_{q_{\theta_e}(\boldsymbol{z}_{\leq T}|\boldsymbol{x}_{\leq T})} \Big[\sum_{t=1}^{T} \big( & -KL(q_{\theta_e}(\boldsymbol{z}_t | \boldsymbol{x}_{\leq t}, \boldsymbol{z}_{< t})\|p(\boldsymbol{z}_t | \boldsymbol{x}_{< t}, \boldsymbol{z}_{< t})) \\& +  \log p_{\theta_d}(\boldsymbol{x}_t|\boldsymbol{x}_{\leq t}, \boldsymbol{z}_{< t}) \big) \Big]
\end{aligned}
\end{small}\label{eq:vrnn_loss}
\end{equation}
where $KL(\cdot\|\cdot)$ refers to Kullback-Leibler divergence, $q_{\theta_e}$ is the inference model, $p$ is the prior, and $p_{\theta_d}$ is the generative model. Note that $\boldsymbol{z}_{\leq T}$ refers to all $\boldsymbol{z}_t$ such that $t\leq T$.
Further, the classification loss is defined based on the binary cross-entropy loss  ($\mathcal{L}_B$)  as:
\begin{equation}
    \mathcal{L}_{\text{clf}}(\boldsymbol{\mathrm{\boldsymbol{x}}};\theta_c, \theta_e)=\mathcal{L}_B(C_{\theta_c}(E(\boldsymbol{\mathrm{x}};\theta_e)),\mathrm{y})
\end{equation}
\noindent where $C(\cdot)$ represents the VRNN classifier with parameters $\theta_c$, and $E(\cdot)$ is the VRNN encoder that maps $\boldsymbol{\mathrm{x}}\rightarrow\tilde{\boldsymbol{z}}$. The above two losses are optimized iteratively until convergence.

\subsection{WGAN for Adversarial Adaptation}\label{sec:gan}
In a traditional Generative Adversarial Network (GAN) \cite{gan2014goodfellow}, generator $G$ and discriminator $D$ play a minimax game where $D$ tries to classify the generated samples as fake and $G$ tries to fool $D$ by producing samples that are as realistic as possible. In Wasserstein GAN (WGAN) \cite{arjovsky2017wasserstein}, instead of binary classification, the discriminator tries to minimize the earth mover distance $W(p,q)$ between real and fake data distributions and the objective function is defined as:
\begin{equation}
   \min_G \max_{D} \mathbb{E}_{\boldsymbol{x}\sim \mathbb{P}_r}[D(\boldsymbol{x})] - \mathbb{E}_{\tilde{\boldsymbol{x}} \sim \mathbb{P}_g}[D(\tilde{\boldsymbol{x}})]
\end{equation}
where $\mathbb{P}_r$ is the real data distribution and $\mathbb{P}_g$ is the model distribution implicitly defined by $\tilde{\boldsymbol{x}}=G(\boldsymbol{z})$, while $\boldsymbol{z} \sim p_z$ being latent random noise. In this case, minimizing the value function with respect to the generator parameters will minimize $W(\mathbb{P}_r, \mathbb{P}_g)$. To improve training stability and efficiency of WGAN, a penalizing term of the gradient’s norm is added to the discriminator's objective function (WGAN-GP)  \cite{gulrajani2017improved}. This will result in the following loss function for $D$ to minimize:
\begin{equation}\label{d_loss}
    \mathcal{L}_\mathrm{dis} (\boldsymbol{x}, \tilde{\boldsymbol{x}};\theta_\mathrm{dis}) = D_{\theta_\mathrm{dis}}(\tilde{\boldsymbol{x}}) - D_{\theta_\mathrm{dis}}(\boldsymbol{x}) + \lambda (\|\nabla_{\hat{\boldsymbol{x}}} D_{\theta_\mathrm{dis}}(\hat{\boldsymbol{x}})\|-1)^2
\end{equation}
 where $\hat{\boldsymbol{x}} = \epsilon \boldsymbol{x} + (1-\epsilon) \tilde{\boldsymbol{x}}$, $\theta_\mathrm{dis}$ indicates the discriminator's parameters, and $\lambda=10$ is the gradient penalty coefficient.

Our proposed framework leverages this functionality of WGAN-GP to minimize the discrepancies between latent representations from each domains  to generate \emph{domain-invariant} features. Also, a unified classifier is learned simultaneously to ensure \emph{class-discriminative} representations.

\subsection{Adversarial Domain Unification (ADU)}\label{sec:adu}
Adversarial Domain Unification (ADU) framework has an asymmetric architecture that differentiates between the two domains, by formulating them as source and target domains: $\mathcal{D}_1=\mathcal{S}$ and  $\mathcal{D}_2=\mathcal{T}$. ADU unifies latent representation of both domains by mapping the target latent representations to align pre-trained source latent representation distribution through adversarial learning. By performing this alignment, ADU accounts for the covariate shift between domains.

\subsubsection{Step 1: Pre-train Source VRNN}
We construct VRNN$_\mathcal{S}$ by following the learning procedure described in Section \ref{vrnn} for the source domain  to find the optimal latent representation, $\boldsymbol{\mathrm{z}}_\mathcal{S}$. The parameters of this network are fixed in the next step to learn a source-like latent representation for target domain.

\subsubsection{Step 2: Adversarial Unification}
\begin{figure}[tb]
\centering
\includegraphics[width=0.47\textwidth]{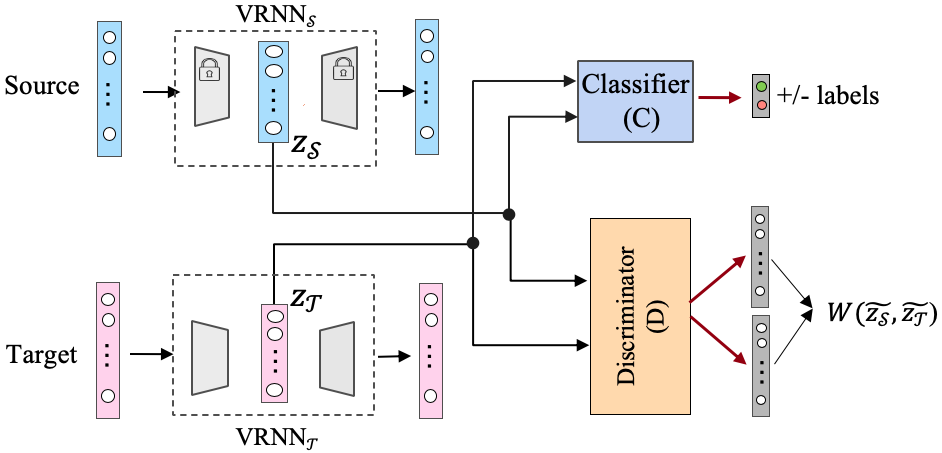}
\caption{\small{Adversarial Domain Unification (ADU)}}
\label{fig:adu}
\end{figure}

Fig. \ref{fig:adu} shows the adversarial unification procedure. Instead of the sequential structure of VRNN, we show it by an encoder-decoder structure. In this step, the target VRNN learns the mapping to align target latent space $\boldsymbol{\mathrm{z}}_\mathcal{T}$ to pre-trained $\boldsymbol{\mathrm{z}}_\mathcal{S}$, using the adversarial discriminator. Further, simultaneous learning of a classifier ensures a class-discriminative mapping. In the following we describe the interaction between components of ADU, the loss function for each, and the overall optimization procedure.

\begin{figure*}[tb]
\centering
\includegraphics[width=0.68\textwidth]{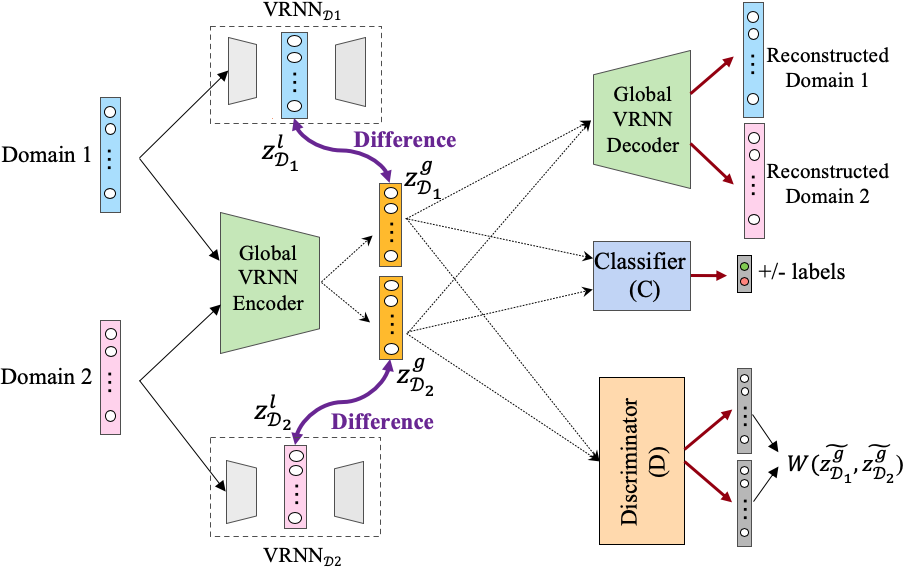}
\caption{VRNN-based Adversarial Domain Separation (VR-ADS)}
\label{fig:ads-step2}
\end{figure*}

    \noindent\textbf{Target VRNN}: VRNN$_\mathcal{T}$ has the same architecture as the pre-trained VRNN$_\mathcal{S}$.
    Since the objective is to align $\boldsymbol{\mathrm{z}}_\mathcal{T}$ with $\boldsymbol{\mathrm{z}}_\mathcal{S}$, we use the parameters of pre-trained source VRNN's decoder for the target VRNN's decoder to generate source-like outputs. Therefore, the parameters of target encoder ($E_\mathcal{T}$), noted as $\theta_{e_t}$, are only updated in this step. The target VRNN loss objective is $\mathcal{L}_{\text{vrnn}_\mathcal{T}}(\boldsymbol{\mathrm{x}}_\mathcal{T};\theta_{e_t})$ as defined by the original VRNN loss illustrated in Eq. \eqref{eq:vrnn_loss}. Further, the encoder $E_\mathcal{T}$ receives feedback from the discriminator to generate invariant latent representation. Therefore, its adversarial loss is defined based on the generator's loss in a WGAN architecture as:
    \begin{equation}
        \mathcal{L}_{\text{adv}}(\boldsymbol{\mathrm{x}}_\mathcal{T};\theta_{e_t}) = D_{\theta_\mathrm{dis}}(E_\mathcal{T}(\boldsymbol{\mathrm{x}}_\mathcal{T});\theta_{e_t})).
    \end{equation}

    The overall unification loss is defined by:
    \begin{equation}
        \mathcal{L}_{\mathrm{unif}}(\boldsymbol{\mathrm{x}}_\mathcal{T};\theta_{e_t}) =   \mathcal{L}_{\text{vrnn}_\mathcal{T}}(\boldsymbol{\mathrm{x}}_\mathcal{T};\theta_{e_t}) + \alpha \mathcal{L}_{\text{adv}}(\boldsymbol{\mathrm{x}}_\mathcal{T};\theta_{e_t})
    \end{equation}
    where $\alpha$ is adversarial loss weight.

    \noindent\textbf{Discriminator}: The discriminator $D$ is an LSTM network that takes the pre-trained temporal   $\boldsymbol{\mathrm{z}}_\mathcal{S}$ and $\boldsymbol{\mathrm{z}}_\mathcal{T}$ as inputs. The output is a vector produced by the last time step of this network. The discriminator loss is defined as a Wasserstein distance between output vectors $W(\widetilde{ \boldsymbol{\mathrm{z}}_\mathcal{S}}, \widetilde{\boldsymbol{\mathrm{z}}_\mathcal{T}})$, and the objective is to maximize this distance for it to be able to distinguish between input sources. The discriminator's loss is $\mathcal{L}_\mathrm{dis}(\boldsymbol{\mathrm{z}}_\mathcal{S}, \boldsymbol{\mathrm{z}}_\mathcal{T};\theta_\mathrm{dis})$ as defined in Eq. \eqref{d_loss}.

    \noindent\textbf{Classifier}: A fully connected neural network is used as a classifier, which takes the latent representation of the last time step, $z_{T}$, as input from both $\boldsymbol{\mathrm{z}}_\mathcal{S}$ and $\boldsymbol{\mathrm{z}}_\mathcal{T}$. This sub-network is optimized based on the binary cross-entropy loss ($\mathcal{L}_B$) in both domains:
    \begin{equation}
    \begin{split}
        \mathcal{L}_{\text{clf}}(\mathcal{D}_\mathcal{S}, \mathcal{D}_\mathcal{T};\theta_c,\theta_{e_t}) =  &
        \mathcal{L}_B(C_{\theta_c}(E_{\mathcal{S}}(\boldsymbol{\mathrm{x}}_\mathcal{S})_{T}), \mathrm{y}_\mathcal{S})\\&+ 
        \mathcal{L}_B(C_{\theta_c}(E_{_\mathcal{T}}(\boldsymbol{\mathrm{x}}_\mathcal{T};\theta_{e_t})_{T}), \mathrm{y}_\mathcal{T})
    \end{split}
    \end{equation}
    where $\theta_c$ indicates the classifier parameters.

All these three components are optimized alternatively based on their objectives until convergence.

In short, following prior DA research, our ADU also aligns latent representations between source and target as much as possible and as a result, we believe it can effectively address the covariate shift. However, ADU does not address the fact that aligning domain-specific information caused by systematic bias would result in contamination in the shared invariant space. Therefore, we propose VR-ADS in the following to address both types of discrepancies.

\section{VRNN-based Adversarial Domain Separation (VR-ADS)}\label{sec:ads}
 We propose VR-ADS framework that separates one globally-shared latent representation for all domains from domain-specific (local) information. This will allow the global information to be purified so that the systematic bias is addressed. VR-ADS pre-trains local VRNNs to generate local latent representations and ensures that the global latent representations are different from the local ones by maximizing a dissimilarity measure. Similar to ADU, a discriminator and a classifier are used to ensure domain-invariant and class-discriminative projection.

\subsubsection{Step 1: Pre-train domain-specific VRNNs}
Optimal local latent representations $\boldsymbol{\mathrm{z}}_{\mathcal{D}_1}^l$ and $\boldsymbol{\mathrm{z}}_{\mathcal{D}_2}^l$ are obtained by pre-training a VRNN classifier for each domain separately, following the learning steps described in Section \ref{vrnn}.

\subsubsection{Step 2: Adversarial Separation}
Fig. \ref{fig:ads-step2} illustrates VR-ADS architecture. This framework is composed of two pre-trained local VRNNs for each domain from step 1, and one global VRNN that takes the concatenation of both domains as input. The global Encoder will generate global latent representations, and the global decoder reconstructs the input for each domain. The discriminator aligns the global representations from $\mathcal{D}_1$ and $\mathcal{D}_2$, and the classifier learns to predict the outcome. Following describes each component's loss.

    \noindent\textbf{Global and local VRNNs}: The parameters of  $\mathrm{VRNN}_{\mathcal{D}_1}$ and $\mathrm{VRNN}_{\mathcal{D}_2}$ are initialized based on the pre-trained VRNNs to generate local representations $\boldsymbol{\mathrm{z}}_{\mathcal{D}_1}^l$ and $\boldsymbol{\mathrm{z}}_{\mathcal{D}_2}^l$. The global VRNN also takes both domain's data as input and the global encoder ($E_g$) generates $\boldsymbol{\mathrm{z}}_{\mathcal{D}_1}^g$ and $\boldsymbol{\mathrm{z}}_{\mathcal{D}_2}^g$. Further, for optimizing reconstruction loss in each of the local and global VRNNs we follow the original VRNN loss defined in Eq. \eqref{eq:vrnn_loss} as follows:
    \begin{equation}
    \begin{split}
    \resizebox{\hsize}{!}{
         $\mathcal{L}^l_\text{vrnn}(\boldsymbol{\mathrm{x}}_{\mathcal{D}_1},\boldsymbol{\mathrm{x}}_{\mathcal{D}_2};\Theta^l) = 
         \mathcal{L}_{\mathrm{vrnn}}(\boldsymbol{\mathrm{x}}_{\mathcal{D}_1};\theta_{e_1},\theta_{d_1}) +
         \mathcal{L}_{\mathrm{vrnn}}(\boldsymbol{\mathrm{x}}_{\mathcal{D}_2};\theta_{e_2},\theta_{d_2})$}
    \end{split}
    \end{equation}
    
    \begin{equation}\label{shared_vrnn_loss}
    \begin{split}
         \mathcal{L}^g_\text{vrnn}(\boldsymbol{\mathrm{x}}_{\mathcal{D}_1},\boldsymbol{\mathrm{x}}_{\mathcal{D}_2};\Theta^g) =  \mathcal{L}_{\mathrm{vrnn}}(\boldsymbol{\mathrm{x}}_{\mathcal{D}_1};\Theta^g) + \mathcal{L}_{\mathrm{vrnn}}(\boldsymbol{\mathrm{x}}_{\mathcal{D}_2};\Theta^g) 
    \end{split}
    \end{equation}
    where $\Theta^l=(\theta_{e_1},\theta_{d_1}, \theta_{e_2},\theta_{d_2})$ and $\Theta^g = (\theta_e^g,\theta_d^g)$ indicate the local and global VRNN parameters, respectively.

    The main objective of VR-ADS is to separate local and global features by maximizing the distance between them so that they extract systematic bias. Therefore,
    we add a dissimilarity measure between $(\boldsymbol{\mathrm{z}}_{\mathcal{D}_1}^g , \boldsymbol{\mathrm{z}}_{\mathcal{D}_1}^l)$ and $(\boldsymbol{\mathrm{z}}_{\mathcal{D}_2}^g , \boldsymbol{\mathrm{z}}_{\mathcal{D}_2}^l)$ for each sample, defined by a Frobenius norm. Frobenius norm measures the orthogonality between global and local representation from each domain, while zero indicates orthogonal vectors. 
    Let us denote matrices  $\boldsymbol{Z}_{\mathcal{D}_1}^g$ and $\boldsymbol{Z}_{\mathcal{D}_2}^g$ as global matrices where row $i$ of each is composed of  $\boldsymbol{\mathrm{z}}_{\mathcal{D}_1}^g$ and $\boldsymbol{\mathrm{z}}_{\mathcal{D}_2}^g$ for sample $i$. Similarly, $\boldsymbol{Z}_{\mathcal{D}_1}^l$ and $\boldsymbol{Z}_{\mathcal{D}_2}^l$ indicate local matrices.
    Therefore, the difference loss is defined as:
    
    \begin{equation}
        \mathcal{L}_\mathrm{diff}(\boldsymbol{\mathrm{x}}_{\mathcal{D}_1}, \boldsymbol{\mathrm{x}}_{\mathcal{D}_2};\Theta^l,\Theta^g) = \norm{{\boldsymbol{\mathrm{Z}}_{\mathcal{D}_1}^g}^\top \boldsymbol{\mathrm{Z}}_{\mathcal{D}_1}^l}_F^2 +
        \norm{{\boldsymbol{\mathrm{Z}}_{\mathcal{D}_2}^g}^\top \boldsymbol{\mathrm{Z}}_{\mathcal{D}_2}^l}_F^2
    \end{equation}
    where $\norm{\cdot}_F^2$ refers to the squared Frobenius norm.
    
    Adversarial loss is defined as the Wasserstein distance between the discriminator-generated vectors by:
    
    \begin{equation}
    \resizebox{\hsize}{!}{
        $\mathcal{L}_\mathrm{adv} (\boldsymbol{\mathrm{x}}_{\mathcal{D}_1}, \boldsymbol{\mathrm{x}}_{\mathcal{D}_2};\theta_e^g) = -D_{\theta_\mathrm{dis}}(E_g(\boldsymbol{\mathrm{x}}_{\mathcal{D}_2});\theta_e^g) + D_{\theta_\mathrm{dis}}(E_g(\boldsymbol{\mathrm{x}}_{\mathcal{D}_1});\theta_e^g)$}
    \end{equation}
    such that minimizing this distance by $E_g$ makes the discriminator confused. Finally, the overall separation loss is:
    
    \begin{equation}\label{sep_loss}
    \begin{split}
        \mathcal{L}_\mathrm{sep}(&\boldsymbol{\mathrm{x}}_{\mathcal{D}_1}, \boldsymbol{\mathrm{x}}_{\mathcal{D}_2};\Theta) =   \resizebox{0.65\hsize}{!}{$\mathcal{L}^l_\text{vrnn}(\boldsymbol{\mathrm{x}}_{\mathcal{D}_1},\boldsymbol{\mathrm{x}}_{\mathcal{D}_2};\Theta^l) + \mathcal{L}^g_\text{vrnn}(\boldsymbol{\mathrm{x}}_{\mathcal{D}_1},\boldsymbol{\mathrm{x}}_{\mathcal{D}_2};\Theta^g) $}
         \\ &  + \alpha
        \mathcal{L}_\mathrm{adv} (\boldsymbol{\mathrm{x}}_{\mathcal{D}_1}, \boldsymbol{\mathrm{x}}_{\mathcal{D}_2};\theta_e^g)
        + \beta \mathcal{L}_\mathrm{diff}(\boldsymbol{\mathrm{x}}_{\mathcal{D}_1}, \boldsymbol{\mathrm{x}}_{\mathcal{D}_2};\Theta^l,\Theta^g).
    \end{split}
    \end{equation}


    \noindent\textbf{Discriminator}: Similar to ADU, the discriminator $D$ is an LSTM network that takes the global representations $\boldsymbol{\mathrm{z}}_{\mathcal{D}_1}^g$ and $\boldsymbol{\mathrm{z}}_{\mathcal{D}_2}^g$ as input. The output is a vector generated by the last time-step of the LSTM for each domain. Its adversarial loss is defined as the Wasserstein distance between discriminator's output vectors,  $W(\widetilde{ \boldsymbol{\mathrm{z}}_{\mathcal{D}_1}^g}, \widetilde{\boldsymbol{\mathrm{z}}_{\mathcal{D}_2}^g})$, and the objective for the discriminator is to maximize this distance so that it would be able to distinguish between input sources. The discriminator's loss is $\mathcal{L}_\mathrm{dis}(\boldsymbol{\mathrm{z}}_{\mathcal{D}_1}^g, \boldsymbol{\mathrm{z}}_{\mathcal{D}_2}^g;\theta_\mathrm{dis})$ as defined by Eq. \eqref{d_loss}.

    \noindent\textbf{Classifier:} A simple fully connected neural network is used as a classifier, which consumes the global latent representations from the last time step $T$. This network is optimized based on the binary cross-entropy loss ($\mathcal{L}_B$) for both domains as:
    
    \begin{equation}
    \begin{split}
        \mathcal{L}_{\text{clf}}(\mathcal{D}_1, \mathcal{D}_2;\theta_c,\theta_e^g) = &
        \mathcal{L}_B(C_{\theta_c}(E_g(\boldsymbol{\mathrm{x}}_{\mathcal{D}_1};\theta_e^g)_{T}), \mathrm{y}_{\mathcal{D}_1})\\&+ 
        \mathcal{L}_B(C_{\theta_c}(E_g(\boldsymbol{\mathrm{x}}_{\mathcal{D}_2};\theta_e^g)_{T}), \mathrm{y}_{\mathcal{D}_2})
    \end{split}
    \end{equation}
    where $\theta_c$ indicates the classifier parameters. 
    
By separating the local and global latent representations enforced by VR-ADS loss functions, we hypothesize that this model will address the systematic bias effectively and outperform ADU for adaptation across EHR systems. We assess this hypothesis through experimentation in the following sections.

\section{Experimental Setup}

\subsection{Dataset Description}
We conduct experiments using two large-scale real-world EHR datasets collected from two U.S. medical systems: Christiana Care Health System (CCHS) and Mayo Clinic. 
To be consistent among both EHRs, we include visits of adult patients (i.e. age$>$18) from July, 2013 to December, 2015. As we discussed in Section \ref{sec:intro}, our preliminary analysis verifies the existence of both covariate shift due to different population demographics and systematic bias due to dramatically different measurement frequency across these two systems.
In total, there are 210,289 visits and more than 10 million medical events in CCHS dataset, and 121,019 visits and more than 53 million events in Mayo dataset. Our study population includes patients with \emph{suspected infection}, which is identified by administration of any anti-infectives, or a positive PCR test result. This population includes 61,848 visits in CCHS and 74,463 visits in Mayo dataset. The definition of study population and the following data pre-processing steps are determined by three leading clinicians from these two healthcare systems with an extensive experience on this subject.

\subsubsection{Labeling} International Classification of Disease (ICD-9) codes are widely used for clinical labeling (i.e., septic shock or not). However, these labels are usually generated for administration purposes at the end of visit and they cannot identify when septic shock happened during a visit. Also, previous studies have demonstrated that such labels are unreliable to be used alone as ground truth \cite{giuliano2007physiological}. Therefore, we refer to a definition from our experts to identify septic shock onset. Following Sepsis-3 definitions \cite{singer2016third}, our clinicians identify septic shock at event-level as having received vasopressor(s) or persistent hypotension for more than 1 hour (systolic blood pressure (SBP)$<$90; or mean arterial pressure$<$65; or drop in SBP$>$40 in an 8-hour window).
\subsubsection{Sampling} We use the intersection of ICD-9 and expert-defined rules to identify the reliable population of septic shock and non-septic shock patients. This results in 2,963 and 3,499 positive cases in CCHS and Mayo, respectively. Since the number of negative cases are much larger than the positives, we perform a stratified random sampling that satisfies two conditions: 1) maintaining the same underlying age, gender, ethnicity, and length of stay distribution, and 2) having the same level of severity as positive samples. The severity of septic shock visits are identified by the presence of initial stages of sepsis in their visit: infection, inflammation, and organ failure as defined by experts.
\subsubsection{Aggregation} Since the sampling frequency of CCHS and Mayo are very different, we consider a 30-minutes aggregation window  to align and regularize time steps in two datasets. 
All measurements within each 30 minutes window are summarized into one event. Our features include 7 vital signs (e.g.: SBP, Temperature), 2 oxygen information (FIO2 and OxygenFlow), and 10 lab results (e.g.: WBC, BUN). For vital signs, we keep min, max, and mean in a window, while for the rest we only keep the mean. If no measurement is taken within an aggregation window, then all values are missing. To handle the missing values in input to our models, we first use expert rules to carry forward vital signs (for 8 hours) and lab results (for 24 hours), then for the rest we apply mean imputation along with adding missing indicator. Our experiments show that this strategy will help VRNN address such variabilities in data more efficiently.

\subsection{Evaluation Task}
\begin{figure}[tb]
\centering
\includegraphics[width=0.48\textwidth]{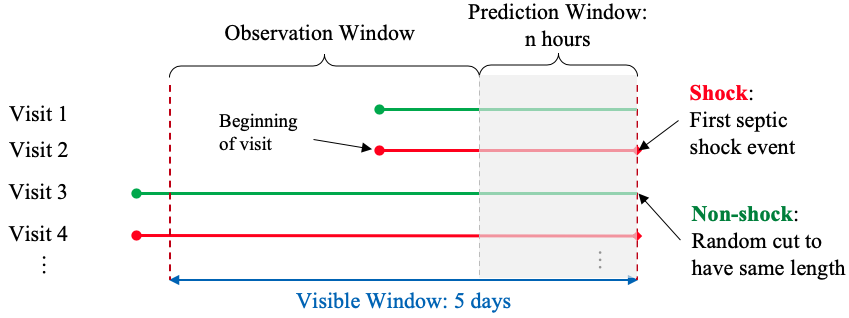}
\caption{\small{Septic shock early prediction task}}\label{fig:early_task}
\end{figure}

Our supervised DA task is to predict whether or not a patient is going to develop septic shock $n$ hours later, given EHRs in observation window. In this work, to minimize the risk of developing septic shock, $n$ varies from 24 to 48 hours as suggested by the leading physicians. To conduct this prediction task, we align all the sequences by their end time, which is the shock onset for positive visits and a truncated time point for non-shock visits. To prevent the potential bias in models, negative visits are truncated such that they have the same distribution of length as positives. Due to the fast progression of sepsis and varying length of sequences, we consider a fixed five days \textit{visible window} prior to the endpoint. The $n$-hours \textit{prediction window} and  \textit{observation window} are illustrated in Fig. \ref{fig:early_task}.  As the prediction window expands, the number of visits remaining in the observation window will drop. For a fair comparison, we sample the same number of positive/negative visits in both domains. This results in 2,132 visits from each domain for 24 hours and 1,530 visits for 48 hours early prediction tasks. Therefore, as the number of samples decrease in these scenarios it is more important to combine different domains through domain adaptation to  build robust classifiers.

\subsection{Baselines and Implementation Details}
 We evaluate the effectiveness of all DA approaches in terms of their ability to accurately predict septic shock in \emph{both} domains. Therefore, the test set is composed of an equal number of CCHS and Mayo test samples. We categorize all models in four groups:

\noindent \textbf{Three Non-adaptive Baselines}:
\begin{enumerate}
    \item[1.] VRNN(CCHS): a VRNN trained on CCHS only.
    \item[2.] VRNN(Mayo): a VRNN trained on Mayo only.
    \item[3.] VRNN(Both): a VRNN trained on  samples from both CCHS and Mayo.
\end{enumerate}

\noindent \textbf{Three Domain Adaptation Baselines} include two variations of a network-based DA and one VRNN-based adversarial DA.
\begin{enumerate}
    \item[4.] FT(C$\rightarrow$M) \cite{alves2018dynamic}: a VRNN pre-trained on CCHS as source domain and fine-tuned (FT) on Mayo as target domain.
    \item[5.] FT(M$\rightarrow$C) \cite{alves2018dynamic}: a VRNN pre-trained on Mayo as source domain and fine-tuned (FT) on CCHS as target domain.
    \item[6.] VRADA \cite{purushotham2016varda}: a shared VRNN with label classifier trained for both domains while invariant representations are generated by an adversarially-trained discriminator.
\end{enumerate}

\noindent \textbf{Two ADU variations:} alternate the source and target domains:
\begin{enumerate}
    \item[7.]ADU(C$\rightarrow$M): unified in an CCHS-like latent space.
    \item[8.] ADU(M$\rightarrow$C): unified in an Mayo-like latent space.
\end{enumerate}

\noindent \textbf{Two VR-ADS variations:}  whether to fix or release the local VRNN parameters. 
\begin{enumerate}
    \item[9.] VR-ADS(fixed): a VR-ADS that eliminates $\mathcal{L}^l_\text{vrnn}$ from the total loss in Eq. \eqref{sep_loss}.
    \item[10.] VR-ADS(released): a VR-ADS that optimizes the local representations with other components based on Eq. \eqref{sep_loss}.
\end{enumerate}

\begin{table*}[t]
\caption{\small{LSTM vs. VRNN: Average Performance ($\pm$ standard deviation) from 24 to 48 hours on CCHS vs. Mayo \textbf{separately}.}}
\begin{adjustbox}{width=0.84\textwidth,center}
\begin{tabular}{c|c|ccccc}
\specialrule{.1em}{.1em}{.1em}
\textbf{Domain}    &   \textbf{Model }  &   \textbf{Accuracy} &  \textbf{Precision}   & \textbf{Recall/Sensitivity} &  \textbf{F-measure}   & \textbf{AUC} \\ \hline
\multirow{2}{*}{CCHS}   &   LSTM    &   0.62($\pm$0.029)	&   0.688($\pm$0.039)	&   0.536($\pm$0.072)	&   0.572($\pm$0.072)	&   0.675($\pm$0.024)\\
                        &   VRNN    &   \textbf{0.72}($\pm$0.024)	&   \textbf{0.739}($\pm$0.039)	&   \textbf{0.686}($\pm$0.056)	&   \textbf{0.708}($\pm$0.056)	&   \textbf{0.764}($\pm$0.032)\\ \hline
\multirow{2}{*}{Mayo}   &   LSTM    &   0.588($\pm$0.034)	&   0.613($\pm$0.065)	&   0.635($\pm$0.171)	&   0.587($\pm$0.171)	&   0.607($\pm$0.026)\\
                        &   VRNN    &   \textbf{0.717}($\pm$0.03)	&   \textbf{0.737}($\pm$0.04)	&   \textbf{0.686}($\pm$0.057)	&   \textbf{0.708}($\pm$0.057)	&   \textbf{0.752}($\pm$0.034)\\
\specialrule{.1em}{.1em}{.1em}
\end{tabular}
\end{adjustbox}
\label{tab:base}
\end{table*}

\begin{table*}[t]
\caption{\small{Ten DA models: Performance ($\pm$ standard deviation) for 48 hours early prediction on \textbf{both CCHS and Mayo}.}}
\begin{adjustbox}{width=0.94\textwidth,center}
\begin{tabular}{l||ccccc}
\specialrule{.15em}{.1em}{.1em}
\textbf{Model }  &   \textbf{Accuracy} &  \textbf{Precision}   & \textbf{Recall/Sensitivity} &  \textbf{F-measure}   & \textbf{AUC} \\ \hline
1. VRNN (CCHS)                  & 0.675($\pm$0.042)       & \textbf{0.712}($\pm$0.060)	    & 0.597($\pm$0.089)	    & 0.646($\pm$0.089)	    & 0.709($\pm$0.049)\\ 
2. VRNN (Mayo)                  & 0.673($\pm$0.041)        & 0.678($\pm$0.062)     & 0.665($\pm$0.056)     & 0.668($\pm$0.056)     & 0.714($\pm$0.036)\\ 
3. VRNN (Both)                  & \textbf{0.700}($\pm$0.015)       & 0.710($\pm$0.024)	    & \textbf{0.682}($\pm$0.019)	  &  \textbf{ 0.695}($\pm$0.019) &\textbf{0.749}($\pm$0.027)\\ \hline
4. FT (C$\rightarrow$M) \cite{alves2018dynamic}& 0.688($\pm$0.024)       & 0.694($\pm$0.045)	    & {\textbf{0.691}}\textsuperscript{\textdaggerdbl}($\pm$0.043)	    & 0.690($\pm$0.043)	    & 0.729($\pm$0.026)\\ 
5. FT (M$\rightarrow$C) \cite{alves2018dynamic}& 0.698($\pm$0.011)       & 0.729($\pm$0.035)	    & 0.640($\pm$0.048)	    & 0.679($\pm$0.048)	    & 0.754($\pm$0.015)\\ 
6. VRADA \cite{purushotham2016varda}   & \textbf{{0.723}\textsuperscript{\textdaggerdbl}}($\pm$0.022)   & \textbf{0.766**}($\pm$0.043)           & 0.647($\pm$0.056)         & \textbf{0.699}($\pm$0.029)         & \textbf{0.757}($\pm$0.024) \\\hline \hline
7. ADU (C$\rightarrow$M)       & 0.700($\pm$0.034)       & 0.718($\pm$0.038)	    & 0.666($\pm$0.078)	    & 0.688($\pm$0.050)	    & 0.742($\pm$0.041)\\ 
8. ADU (M$\rightarrow$C)       & \textbf{0.713}($\pm$0.035)       & \textbf{0.736}($\pm$0.065)	    & \textbf{0.674}($\pm$0.065)	    & \textbf{0.699}($\pm$0.032)	    & \textbf{0.749}($\pm$0.041)\\ \hline
9. VR-ADS (fixed)                   & 0.720($\pm$0.050)       & 0.742($\pm$0.063)	    & 0.675($\pm$0.078)	    & {0.704}\textsuperscript{\textdaggerdbl}($\pm$0.054)	    & {0.766}\textsuperscript{\textdaggerdbl}($\pm$0.058)\\ 
10. VR-ADS (released)                & \textbf{0.741**}($\pm$0.040)       & {\textbf{0.747}}\textsuperscript{\textdaggerdbl}($\pm$0.056)	    & \textbf{0.732**}($\pm$0.045)	    & \textbf{0.737**}($\pm$0.034)	    & \textbf{0.793**}($\pm$0.047)\\
\specialrule{.15em}{.1em}{.1em}
\end{tabular}
\end{adjustbox}
\label{tab:res-48all}
\begin{tablenotes}
\small 
    \item \hspace{5mm} $\cdot$ For each block, the best model is in \textbf{bold}; The \emph{best} and the \emph{second-best} models across ALL are labeled with $**$ and \textsuperscript{\textdaggerdbl}, respectively.
\end{tablenotes}
\end{table*}

All models are implemented in Tensorflow with the same experimental setup.  The VRNN hidden size and latent size are set as 30 and 50, respectively. The LSTM-based discriminator has depth of one, with 60 hidden units. The batch size is 32 and all models are trained in 160 epochs with early stopping. Validation loss is measured every 10 epochs to choose the best model. The sequences are zero padded to have the same length and the zero-padding is masked for reconstruction loss calculation. 
Three sets of parameters: discriminator ($\theta_d$), classifier ($\theta_c$), and VRNN-based parameters are optimized independently by NAdam optimizer \cite{tato2018nadam}, with learning rates: $\alpha_d=10^{-5}$, $\alpha_c=0.006$, and $\alpha_v=0.003$. During training, we update the above three parameters by alternately optimizing their objectives. 
In every epoch, the order of optimization between the three optimizer are altered to prevent over-training of a specific sub-network. 

\noindent \textbf{Evaluation Metrics and Cross Validation}:
Our evaluation metrics include Accuracy, Precision, Recall/Sensitivity, F-measure, and AUC to include both important metrics in machine learning research (e.g.: AUC, F-measure), and the ones commonly used in healthcare research (e.g.: Precision, Recall). All our experiments are evaluated using 2-fold cross-validation, and the mean and standard deviation from three independent runs are reported.

\section{Results}

\subsection{VRNN vs. LSTM}\label{sec:base_res}
To determine the best base classifier, a comparison between VRNN and LSTM (with depth size 2 and 60 hidden units) was conducted on CCHS and Mayo separately. 
Table \ref{tab:base} shows the average performance of VRNN vs. LSTM on the task of  early prediction of septic shock by varying early prediction window from 24 to 48 hours before the onset, with every 4 hours interval. In this table, the first column shows \emph{the domain that the corresponding model is trained and evaluated}. Table \ref{tab:base}  shows that VRNN significantly outperforms LSTM across both CCHS and Mayo datasets. We hypothesize that this difference is due to the high missing rate in EHR data that can be effectively handled by a VRNN but not LSTM. In the following, all our DA models are based on VRNNs.

\subsection{Domain Adaptation Approaches}
 We will first compare the ten DA models on the most challenging task of early prediction of  septic shock: 48 hours before the onset, and then varying from 24 to 48 hours prior to onset, with every 4 hours interval.

\noindent\textbf{48 hours Early Prediction:}  Table \ref{tab:res-48all} compares the 10 DA models' performance by dividing them into four blocks: Non-adaptive, DA baselines, ADU, and VR-ADS. 
The top block compares the three non-adaptive baselines. It shows that by leveraging information in both domains, VRNN(Both) outperforms the other two for  all metrics except precision. Note that our test data for this experiment is a combination of CCHS and Mayo EHRs. Therefore, the fact that VRNN(Both) outperforms VRNN(CCHS) and VRNN(Mayo) suggests that a model trained on one medical system may not be generalizable to another system effectively.
The second block shows that among the three  DA baselines,   VRADA outperforms all the others across all metrics except for recall. Among network-based methods, FT(M$\rightarrow$C) outperforms FT(C$\rightarrow$M) for all metrics except for recall and F-measure. The third block of Table \ref{tab:res-48all} shows that ADU(M$\rightarrow$C) outperforms ADU(C$\rightarrow$M) for all metrics that is consistent with FT results. Also, ADU(M$\rightarrow$C) performing highly similar to VRNN(Both) indicates that forcing to unify domains might still carry systematic bias to the shared representation that results in inefficient adaptation. 

\renewcommand{\arraystretch}{1.0}
\begin{table*}[t]
\caption{\small{Average Performance ($\pm$ standard deviation) of DA baselines and the best two DA models from  24 to 48 hours. }}
\begin{adjustbox}{width=0.94\textwidth,center}
\begin{tabular}{l||ccccc}
\specialrule{.15em}{.1em}{.1em}
\textbf{Model }  &   \textbf{Accuracy} &  \textbf{Precision}   & \textbf{Recall/Sensitivity} &  \textbf{F-measure}   & \textbf{AUC} \\ \hline
1. VRNN (CCHS)	                & 0.694($\pm$0.018) &	0.713($\pm$0.032) &	0.658($\pm$0.061)   &	0.681($\pm$0.061)   &	0.734($\pm$0.028)\\ 
2. VRNN (Mayo)	                & 0.703($\pm$0.026) &   0.72($\pm$0.039)  &	0.674($\pm$0.053)   &	0.693($\pm$0.053)   &	0.743($\pm$0.029)\\
3. VRNN (Both)	                & \textbf{0.713}($\pm$0.019) &	\textbf{0.732}($\pm$0.034) &	\textbf{0.679}($\pm$0.054)   &	\textbf{0.702}($\pm$0.051)   &	\textbf{0.755}($\pm$0.031)\\ \hline
4. FT (C$\rightarrow$M)	\cite{alves2018dynamic}& 0.701($\pm$0.031) &	0.711($\pm$0.039) &	0.682($\pm$0.056)   &	0.694($\pm$0.056)   &	0.741($\pm$0.039)\\ 
5. FT (M$\rightarrow$C)	\cite{alves2018dynamic}& 0.716($\pm$0.019) &	0.734($\pm$0.033) &	\textbf{0.684}($\pm$0.053)   &   0.705($\pm$0.051)   &	0.763($\pm$0.030)\\
6. VRADA \cite{purushotham2016varda} & \textbf{0.728}($\pm$0.015) &	{\textbf{0.753}}\textsuperscript{\textdaggerdbl}($\pm$0.025) &	0.683($\pm$0.039) &	\textbf{0.714}($\pm$0.019) &	\textbf{0.773}($\pm$0.025) \\ \hline \hline
8. ADU (M$\rightarrow$C)	    & {0.731}\textsuperscript{\textdaggerdbl}($\pm$0.031) &	0.751($\pm$0.043) &	{0.699}\textsuperscript{\textdaggerdbl}($\pm$0.058)   &	{0.721}\textsuperscript{\textdaggerdbl}($\pm$0.035)   &	{0.775}\textsuperscript{\textdaggerdbl}($\pm$0.037)\\
10. VR-ADS (released)	            & \textbf{0.741**}($\pm$0.028) &	\textbf{0.757**}($\pm$0.043) &	\textbf{0.716**}($\pm$0.057)   &	\textbf{0.733**}($\pm$0.032)   &	\textbf{0.787**}($\pm$0.036)\\
\specialrule{.15em}{.1em}{.1em}
\end{tabular}
\end{adjustbox}
\label{tab:res-avg}
\begin{tablenotes}
\small
    \item \hspace{5mm} $\cdot$ For each block, the best model is in \textbf{bold}; The \emph{best} and the \emph{second best} models across ALL are labeled with ** and \textsuperscript{\textdaggerdbl}, respectively. 
\end{tablenotes}
\end{table*}

\begin{figure*}[t!]
\centering
    \begin{subfigure}[t]{0.48\textwidth}
    \begin{center}
        \includegraphics[width=\textwidth]{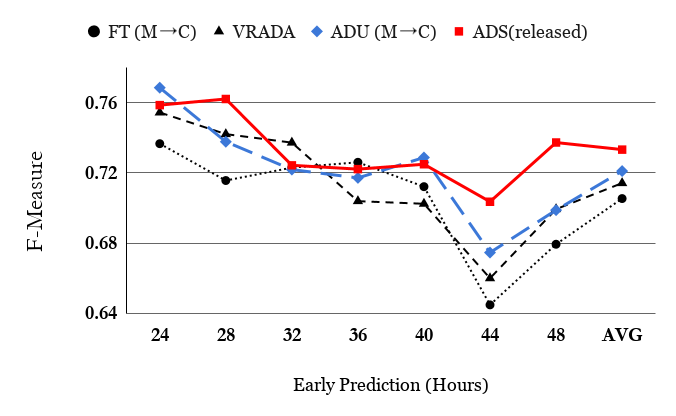}
    \end{center}
    \end{subfigure}
    \begin{subfigure}[t]{0.48\textwidth}
    \begin{center}
        \includegraphics[width=\textwidth]{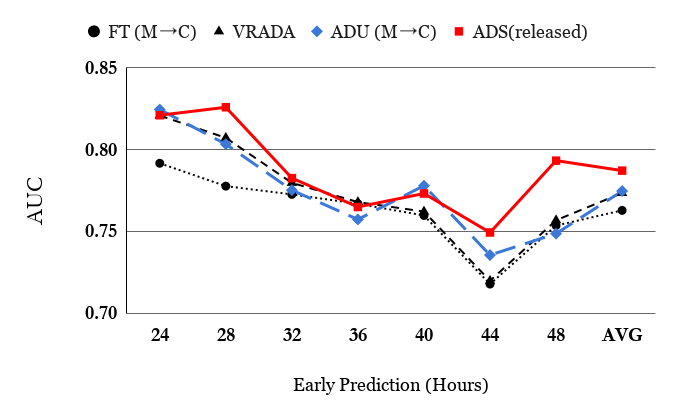}
    \end{center}
    \end{subfigure}
    \caption{\small{Performance comparison across all DA models for 24 to 48 hours early prediction task. The last x-axis point shows the average.}}\label{avg-plot}
\end{figure*}

The bottom block shows that our VR-ADS(released) outperforms VR-ADS(fixed), ADU, and all other baselines for all metrics except for precision. This suggests that it is important to separate the local representation from global representations to capture systematic bias in each domain and generate more robust domain-invariant global representations. The out-performance of VR-ADS(released) over VR-ADS(fixed) shows that as we detach the local characteristics dynamically, the generalization ability of VR-ADS improves in both domains. 
Finally, our results suggest that our proposed DA approaches can address the insufficient labeled data issue: in a separate experiment, we evaluate the performance of VRNN that is trained and tested within the same system and we got a recall/sensitivity of 0.627 and 0.652 for CCHS and Mayo, respectively. The two individual VRNN models would achieve an average recall/sensitivity of 0.639. Table \ref{tab:res-48all}  shows that our VR-ADS(released) framework can perform with 0.732 sensitivity across the two systems, which is  more than 9\% improvement from using the two individually trained models. 

\noindent\textbf{Varying 24-48 hours Early Prediction:}  Table  \ref{tab:res-avg} compares the average performance of best performing variation of our two proposed frameworks, ADU(M$\rightarrow$C) and VR-ADS(released) against all the baselines when varying early prediction window from 24 to 48 hours.  It shows that  VR-ADS(released) outperforms ADU and all baselines across all metrics. This gap is especially significant for the recall/sensitivity with 3.2\% improvement from the best baseline model in the second block.  Most of our findings for 48 hours early prediction still hold for averaging 24 to 48 hours, except that ADU(M$\rightarrow$C) performs slightly better than VRADA and becomes the second-best model. 

Finally, Fig. \ref{avg-plot} compares our two best proposed models, ADU(M$\rightarrow$C) and VR-ADS (released) against the the  two best baselines: FT(M$\rightarrow$C) and VRADA on F-measure and AUC metrics. For F-measure, either ADU or VR-ADS (released) perform better than the  two best baselines on most hours except that they performed slightly worse on 32 and 36 hours early prediction; for AUC, our two proposed models outperform both of the best baselines across all hours. 

\begin{figure}[b!]
\centering
\includegraphics[width=0.47\textwidth]{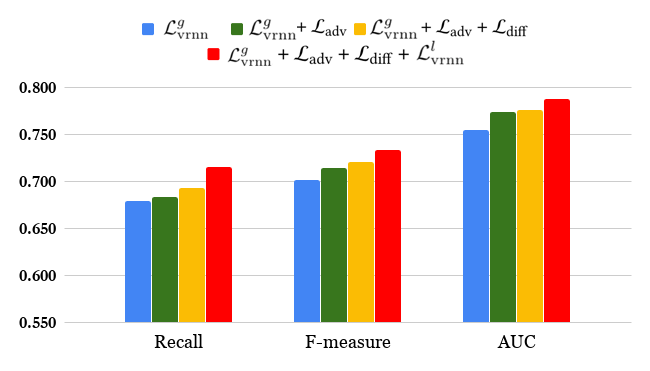}
\caption{\small{Ablation Study for 24-48 hours early prediction. Each column refers to one of the introduced methods from left (blue) to right (red): VRNN(Both), VRADA, VR-ADS(fixed), and VR-ADS(released).
}}
\label{fig:ablation}
\end{figure}

\subsection{Ablation Study}
We conduct an ablation study to evaluate the benefit brought by each loss component introduced in proposed VR-ADS model (Eq. \eqref{sep_loss}). This will assess the robustness of domain-invariant global representation generated in each model. Each loss ablation will result in one of the introduced models. Starting with VRNN(Both), this model only optimizes the global VRNN loss on both domains using $\mathcal{L}^g_\mathrm{vrnn}$ as defined by Eq. \eqref{shared_vrnn_loss}. VRADA is a shared VRNN that is trained adversarially and optimizes $\mathcal{L}^g_\mathrm{vrnn} + \mathcal{L}_\mathrm{adv}$. Further, VR-ADS(fixed) adds the difference loss to address systematic bias and optimizes $\mathcal{L}^g_\mathrm{vrnn} + \mathcal{L}_\mathrm{adv} + \mathcal{L}_\mathrm{diff}$. Finally, our VR-ADS(released) model simultaneously trains the local VRNNs to dynamically extract the biases and optimizes $\mathcal{L}^g_\mathrm{vrnn} + \mathcal{L}_\mathrm{adv} + \mathcal{L}_\mathrm{diff} + \mathcal{L}^l_\mathrm{vrnn}$. Recall, F-measure, and AUC performance of these four models are compared in Fig. \ref{fig:ablation}. We find that the performance increases by progressively adding loss components, such that the VR-ADS(released) achieves the best. This result verifies the importance of domain adaptation by addressing both covariate shift and systematic bias as embedded in VR-ADS framework.

\section{Conclusion}
Developing a robust, generalizable model for septic shock early prediction is a crucial yet challenging task. In this work, we demonstrate the effectiveness of variational recurrent neural network (VRNN) for septic shock early prediction. Further, we propose a general VRNN-based domain separation framework, called VR-ADS, to address two types of discrepancies across EHR systems due to heterogeneous patient populations (covariate shift) and incompatible data collection procedures (systematic bias). Through experiments, we show that VR-ADS outperforms state-of-the-art DA methods because of its ability to fully extract local bias from the global representations, and addressing both covariate shift and systematic bias. In the future, we will leverage the robustness and symmetric architecture of VR-ADS and investigate its application to multi-domain adaptation (unifying more than two systems) and domain generalization scenarios (adaptation to any unseen system). \\
 
\noindent{\textbf{Acknowledgement:}}
This research was supported by the NSF Grants 2013502, 1726550, 1651909, and 1522107.

\bibliographystyle{IEEEtran}
\bibliography{IEEEabrv,IEEE_BigData}

\begin{thebibliography}{10}
\providecommand{\url}[1]{#1}
\csname url@samestyle\endcsname
\providecommand{\newblock}{\relax}
\providecommand{\bibinfo}[2]{#2}
\providecommand{\BIBentrySTDinterwordspacing}{\spaceskip=0pt\relax}
\providecommand{\BIBentryALTinterwordstretchfactor}{4}
\providecommand{\BIBentryALTinterwordspacing}{\spaceskip=\fontdimen2\font plus
\BIBentryALTinterwordstretchfactor\fontdimen3\font minus
  \fontdimen4\font\relax}
\providecommand{\BIBforeignlanguage}[2]{{%
\expandafter\ifx\csname l@#1\endcsname\relax
\typeout{** WARNING: IEEEtran.bst: No hyphenation pattern has been}%
\typeout{** loaded for the language `#1'. Using the pattern for}%
\typeout{** the default language instead.}%
\else
\language=\csname l@#1\endcsname
\fi
#2}}
\providecommand{\BIBdecl}{\relax}
\BIBdecl

\bibitem{agniel2018biases}
D.~Agniel, I.~S. Kohane, and G.~M. Weber, ``Biases in electronic health record
  data due to processes within the healthcare system: retrospective
  observational study,'' \emph{Bmj}, vol. 361, 2018.

\bibitem{alves2018dynamic}
T.~Alves, A.~Laender, A.~Veloso, and N.~Ziviani, ``Dynamic prediction of icu
  mortality risk using domain adaptation,'' in \emph{2018 IEEE International
  Conference on Big Data (Big Data)}.\hskip 1em plus 0.5em minus 0.4em\relax
  IEEE, 2018, pp. 1328--1336.

\bibitem{purushotham2016varda}
S.~Purushotham, W.~Carvalho, T.~Nilanon, and Y.~Liu, ``Variational recurrent
  adversarial deep domain adaptation,'' 2016.

\bibitem{singer2016third}
M.~Singer, C.~S. Deutschman, C.~W. Seymour, M.~Shankar-Hari \emph{et~al.},
  ``The third international consensus definitions for sepsis and septic shock
  (sepsis-3),'' \emph{Jama}, vol. 315, no.~8, pp. 801--810, 2016.

\bibitem{cdc2016}
\BIBentryALTinterwordspacing
C.~for Disease~Control and Prevention, ``National center for emerging and
  zoonotic infectious diseases (ncezid), division of healthcare quality
  promotion (dhqp),'' August 23, 2016. [Online]. Available:
  \url{https://www.cdc.gov/sepsis/datareports/index.html}
\BIBentrySTDinterwordspacing

\bibitem{torio2006national}
C.~Torio and R.~Andrews, ``National inpatient hospital costs: the most
  expensive conditions by payer, 2011: statistical brief\# 160,'' 2006.

\bibitem{kumar2006duration}
A.~Kumar, D.~Roberts, K.~E. Wood, B.~Light, J.~E. Parrillo, S.~Sharma,
  R.~Suppes, D.~Feinstein, S.~Zanotti, L.~Taiberg \emph{et~al.}, ``Duration of
  hypotension before initiation of effective antimicrobial therapy is the
  critical determinant of survival in human septic shock,'' \emph{Critical care
  medicine}, vol.~34, no.~6, pp. 1589--1596, 2006.

\bibitem{macdonald2014comparison}
S.~P. Macdonald, G.~Arendts, D.~M. Fatovich, and S.~G. Brown, ``Comparison of
  piro, sofa, and meds scores for predicting mortality in emergency department
  patients with severe sepsis and septic shock,'' \emph{Academic Emergency
  Medicine}, vol.~21, no.~11, pp. 1257--1263, 2014.

\bibitem{henry2015targeted}
K.~E. Henry, D.~N. Hager, P.~J. Pronovost, and S.~Saria, ``A targeted real-time
  early warning score (trewscore) for septic shock,'' \emph{Science
  translational medicine}, vol.~7, no. 299, pp. 299ra122--299ra122, 2015.

\bibitem{dorsett2017qsofa}
M.~Dorsett, M.~Kroll, C.~S. Smith, P.~Asaro, S.~Y. Liang, and H.~P. Moy,
  ``qsofa has poor sensitivity for prehospital identification of severe sepsis
  and septic shock,'' \emph{Prehospital emergency care}, vol.~21, no.~4, pp.
  489--497, 2017.

\bibitem{mak2019prospective}
M.~H.~W. Mak, J.~K. Low, S.~P. Junnarkar, T.~C.~W. Huey, and V.~G. Shelat, ``A
  prospective validation of sepsis-3 guidelines in acute hepatobiliary sepsis:
  qsofa lacks sensitivity and sirs criteria lacks specificity (cohort study),''
  \emph{International Journal of Surgery}, vol.~72, 2019.

\bibitem{tintinalli2015sepsis}
J.~Tintinalli, S.~J, J.~M. O, C.~D, C.~R, and M.~G, \emph{Tintinallis emergency
  medicine A comprehensive study guide}, 7th~ed.\hskip 1em plus 0.5em minus
  0.4em\relax McGraw-Hill Education, 2011, ch. 146: Septic Shock, pp.
  1003--1014.

\bibitem{lin2019multi}
C.~Lin, J.~Ivy, and M.~Chi, ``Multi-layer facial representation learning for
  early prediction of septic shock,'' in \emph{2019 IEEE International
  Conference on Big Data}.\hskip 1em plus 0.5em minus 0.4em\relax IEEE, 2019,
  pp. 840--849.

\bibitem{lin2018early}
C.~Lin, Y.~Zhangy, J.~Ivy, M.~Capan, R.~Arnold, J.~M. Huddleston, and M.~Chi,
  ``Early diagnosis and prediction of sepsis shock by combining static and
  dynamic information using convolutional-lstm,'' in \emph{2018 IEEE
  ICHI}.\hskip 1em plus 0.5em minus 0.4em\relax IEEE, 2018, pp. 219--228.

\bibitem{zhang2017lstm}
Y.~Zhang, C.~Lin, M.~Chi, J.~Ivy, M.~Capan, and J.~M. Huddleston, ``Lstm for
  septic shock: Adding unreliable labels to reliable predictions,'' in
  \emph{IEEE International Conference on Big Data}, 2017, pp. 1233--1242.

\bibitem{kim2018temporal}
Y.~J. Kim and M.~Chi, ``Temporal belief memory: imputing missing data during
  rnn training,'' in \emph{Proceedings of the 27th International Joint
  Conference on Artificial Intelligence}, 2018, pp. 2326--2332.

\bibitem{khoshnevisan2018recent}
F.~Khoshnevisan, J.~Ivy, M.~Capan, R.~Arnold, J.~Huddleston, and M.~Chi,
  ``Recent temporal pattern mining for septic shock early prediction,'' in
  \emph{2018 IEEE ICHI}.\hskip 1em plus 0.5em minus 0.4em\relax IEEE, 2018, pp.
  229--240.

\bibitem{chung2015recurrent}
J.~Chung, K.~Kastner, L.~Dinh, K.~Goel, A.~C. Courville, and Y.~Bengio, ``A
  recurrent latent variable model for sequential data,'' in \emph{Advances in
  neural information processing systems}, 2015, pp. 2980--2988.

\bibitem{zhang2017medical}
S.~Zhang, P.~Xie, D.~Wang, and E.~P. Xing, ``Medical diagnosis from laboratory
  tests by combining generative and discriminative learning,'' \emph{arXiv
  preprint arXiv:1711.04329}, 2017.

\bibitem{mulyadi2020uncertainty}
A.~W. Mulyadi, E.~Jun, and H.-I. Suk, ``Uncertainty-aware variational-recurrent
  imputation network for clinical time series,'' \emph{arXiv preprint
  arXiv:2003.00662}, 2020.

\bibitem{shavdia2007septic}
D.~Shavdia, ``Septic shock: Providing early warnings through multivariate
  logistic regression models,'' Ph.D. dissertation, Massachusetts Institute of
  Technology, 2007.

\bibitem{ghosh2017septic}
S.~Ghosh, J.~Li, L.~Cao, and K.~Ramamohanarao, ``Septic shock prediction for
  icu patients via coupled hmm walking on sequential contrast patterns,''
  \emph{Journal of biomedical informatics}, vol.~66, 2017.

\bibitem{mao2019one}
Y.~Mao, R.~Zhi, F.~Khoshnevisan, T.~W. Price, T.~Barnes, and M.~Chi, ``One
  minute is enough: Early prediction of student success and event-level
  difficulty during a novice programming task.'' \emph{International
  Educational Data Mining Society}, 2019.

\bibitem{zhang2019attain}
Y.~Zhang, X.~Yang, J.~Ivy, and M.~Chi, ``Attain: attention-based time-aware
  lstm networks for disease progression modeling,'' in \emph{Proceedings of the
  28th IJCAI}.\hskip 1em plus 0.5em minus 0.4em\relax AAAI Press, 2019, pp.
  4369--4375.

\bibitem{yang2018time}
X.~Yang, Y.~Zhang, and M.~Chi, ``Time-aware subgroup matrix decomposition:
  Imputing missing data using forecasting events,'' in \emph{2018 IEEE
  International Conference on Big Data}.\hskip 1em plus 0.5em minus 0.4em\relax
  IEEE, 2018, pp. 1524--1533.

\bibitem{chien2017variational}
J.-T. Chien, K.-T. Kuo \emph{et~al.}, ``Variational recurrent neural networks
  for speech separation,'' in \emph{Annual Conference of the International
  Speech Communication Association (InterSpeech)}, 2017, pp. 1193--1197.

\bibitem{pineau2019variational}
E.~Pineau and N.~de~Lara, ``Variational recurrent neural networks for graph
  classification,'' in \emph{Representation Learning on Graphs and Manifolds
  Workshop}, 2019.

\bibitem{tzeng2017adda}
E.~Tzeng, J.~Hoffman, K.~Saenko, and T.~Darrell, ``Adversarial discriminative
  domain adaptation,'' in \emph{Proceedings of the IEEE CVPR}, 2017, pp.
  7167--7176.

\bibitem{pan2010domain}
S.~J. Pan, I.~W. Tsang, J.~T. Kwok, and Q.~Yang, ``Domain adaptation via
  transfer component analysis,'' \emph{IEEE Transactions on Neural Networks},
  vol.~22, no.~2, pp. 199--210, 2010.

\bibitem{long2015learning}
M.~Long, Y.~Cao, J.~Wang, and M.~Jordan, ``Learning transferable features with
  deep adaptation networks,'' in \emph{International conference on machine
  learning}, 2015, pp. 97--105.

\bibitem{tzeng2015simultaneous}
E.~Tzeng, J.~Hoffman, T.~Darrell, and K.~Saenko, ``Simultaneous deep transfer
  across domains and tasks,'' in \emph{Proceedings of the IEEE International
  Conference on Computer Vision}, 2015, pp. 4068--4076.

\bibitem{yao2010boosting}
Y.~Yao and G.~Doretto, ``Boosting for transfer learning with multiple
  sources,'' in \emph{2010 IEEE Computer Society Conference on Computer Vision
  and Pattern Recognition}.\hskip 1em plus 0.5em minus 0.4em\relax IEEE, 2010,
  pp. 1855--1862.

\bibitem{fernando2013unsupervised}
B.~Fernando, A.~Habrard, M.~Sebban, and T.~Tuytelaars, ``Unsupervised visual
  domain adaptation using subspace alignment,'' in \emph{Proceedings of the
  IEEE international conference on computer vision}, 2013.

\bibitem{wang2018deep}
M.~Wang and W.~Deng, ``Deep visual domain adaptation: A survey,''
  \emph{Neurocomputing}, vol. 312, pp. 135--153, 2018.

\bibitem{gan2014goodfellow}
I.~Goodfellow, J.~Pouget-Abadie, M.~Mirza, B.~Xu, D.~Warde-Farley, S.~Ozair,
  A.~Courville, and Y.~Bengio, ``Generative adversarial nets,'' in
  \emph{NeurIPS}, 2014, pp. 2672--2680.

\bibitem{hoffman2017cycada}
J.~Hoffman, E.~Tzeng, T.~Park, J.-Y. Zhu, P.~Isola, K.~Saenko, A.~A. Efros, and
  T.~Darrell, ``Cycada: Cycle-consistent adversarial domain adaptation,''
  \emph{arXiv preprint arXiv:1711.03213}, 2017.

\bibitem{ganin2014unsupervised}
Y.~Ganin and V.~Lempitsky, ``Unsupervised domain adaptation by
  backpropagation,'' \emph{arXiv preprint arXiv:1409.7495}, 2014.

\bibitem{shen2018wasserstein}
J.~Shen, Y.~Qu, W.~Zhang, and Y.~Yu, ``Wasserstein distance guided
  representation learning for domain adaptation,'' in \emph{Thirty-Second AAAI
  Conference on Artificial Intelligence}, 2018.

\bibitem{zhang2019time}
Y.~Zhang, X.~Yang, J.~Ivy, and M.~Chi, ``Time-aware adversarial networks for
  adapting disease progression modeling,'' in \emph{2019 IEEE ICHI}.\hskip 1em
  plus 0.5em minus 0.4em\relax IEEE, 2019, pp. 1--11.

\bibitem{wiens2014study}
J.~Wiens, J.~Guttag, and E.~Horvitz, ``A study in transfer learning: leveraging
  data from multiple hospitals to enhance hospital-specific predictions,''
  \emph{Journal of the American Medical Informatics Association}, vol.~21,
  no.~4, pp. 699--706, 2014.

\bibitem{desautels2017using}
T.~Desautels, J.~Calvert, J.~Hoffman, Q.~Mao, M.~Jay, G.~Fletcher, C.~Barton,
  U.~Chettipally, Y.~Kerem, and R.~Das, ``Using transfer learning for improved
  mortality prediction in a data-scarce hospital setting,'' \emph{Biomedical
  informatics insights}, vol.~9, p. 1178222617712994, 2017.

\bibitem{curth2019transferring}
A.~Curth, P.~Thoral, W.~van~den Wildenberg, P.~Bijlstra, D.~de~Bruin,
  P.~Elbers, and M.~Fornasa, ``Transferring clinical prediction models across
  hospitals and electronic health record systems,'' in \emph{Joint European
  Conference on Machine Learning and Knowledge Discovery in Databases}.\hskip
  1em plus 0.5em minus 0.4em\relax Springer, 2019, pp. 605--621.

\bibitem{mhasawade2019population}
V.~Mhasawade, N.~A. Rehman, and R.~Chunara, ``Population-aware hierarchical
  bayesian domain adaptation via multiple-component invariant learning,''
  \emph{arXiv preprint arXiv:1908.09222}, 2019.

\bibitem{arjovsky2017wasserstein}
M.~Arjovsky, S.~Chintala, and L.~Bottou, ``Wasserstein gan,'' \emph{arXiv
  preprint arXiv:1701.07875}, 2017.

\bibitem{gulrajani2017improved}
I.~Gulrajani, F.~Ahmed, M.~Arjovsky, V.~Dumoulin, and A.~C. Courville,
  ``Improved training of wasserstein gans,'' in \emph{NeurIPS}, 2017.

\bibitem{giuliano2007physiological}
K.~K. Giuliano, ``Physiological monitoring for critically ill patients: testing
  a predictive model for the early detection of sepsis,'' \emph{American
  Journal of Critical Care}, vol.~16, no.~2, pp. 122--130, 2007.

\bibitem{tato2018nadam}
A.~Tato and R.~Nkambou, ``Improving adam optimizer,'' 2018.

\end{thebibliography}

\end{document}